\documentclass{nature}

\title{Building fast Bayesian computing machines out of intentionally stochastic, digital parts.}

\author{Vikash Mansinghka$^{1,2,3}$ and Eric Jonas$^{1,3}$}

\usepackage{graphicx}
\begin{document}

\maketitle

\begin{affiliations}
 \item The authors contributed equally to this work.
 \item Computer Science \& Artificial Intelligence Laboratory, MIT
 \item Department of Brain \& Cognitive Sciences, MIT
\end{affiliations}

\begin{abstract}
The brain interprets ambiguous sensory information faster and more
reliably than modern computers, using neurons that are slower and less
reliable than logic gates. But Bayesian inference, which underpins
many computational models of perception and cognition, appears
computationally challenging even given modern transistor speeds and
energy budgets. The computational principles and structures needed to
narrow this gap are unknown. Here we show how to build fast Bayesian
computing machines using intentionally stochastic, digital parts,
narrowing this efficiency gap by multiple orders of magnitude. We find
that by connecting stochastic digital components according to simple
mathematical rules, one can build massively parallel, low precision
circuits that solve Bayesian inference problems and are compatible
with the Poisson firing statistics of cortical neurons. We evaluate
circuits for depth and motion perception, perceptual learning and
causal reasoning, each performing inference over 10,000+ latent
variables in real time --- a 1,000x speed advantage over commodity
microprocessors. These results suggest a new role for randomness in
the engineering and reverse-engineering of intelligent computation.
\end{abstract}

Our ability to see, think and act all depend on our mind's ability to
process uncertain information and identify probable explanations for
inherently ambiguous data. Many computational models of the perception
of motion\cite{weissmotion}, motor learning\cite{kordinbayesian},
higher-level cognition\cite{griffithsoptimal, blaisdellcausal} and
cognitive development\cite{tenenbaumgrow} are based on Bayesian
inference in rich, flexible probabilistic models of the world. Machine
intelligence systems, including Watson\cite{ferruccibuilding},
autonomous vehicles\cite{thrunprobabilisticarticle} and other robots\cite{thrunprobabilisticbook} and the Kinect\cite{shottonreal} system
for gestural control of video games, also all depend on probabilistic
inference to resolve ambiguities in their sensory input. But brains
solve these problems with greater speed than modern computers, using
information processing units that are orders of magnitude slower and
less reliable than the switching elements in the earliest electronic
computers. The original UNIVAC I ran at 2.25 MHz\cite{univac}, and RAM
from twenty years ago had one bit error per 256 MB per month\cite{shivakumar2002modeling}. In contrast, the fastest neurons in
human brains operate at less than 1 kHz, and synaptic transmission can
completely fail up to 50\% of the time\cite{synapticfailure}.

This efficiency gap presents a fundamental challenge for computer
science. How is it possible to solve problems of probabilistic
inference with an efficiency that begins to approach that of the
brain? Here we introduce intentionally stochastic but still digital
circuit elements, along with composition laws and design rules, that
together narrow the efficiency gap by multiple orders of magnitude.

Our approach both builds on and departs from the principles behind
digital logic. Like traditional digital gates, stochastic digital
gates consume and produce discrete symbols, which can be represented
via binary numbers. Also like digital logic gates, our circuit
elements can be composed and abstracted via simple mathematical rules,
yielding larger computational units that whose behavior can be
analyzed in terms of their constituents. We describe primitives and
design rules for both stateless and synchronously clocked
circuits. But unlike digital gates and circuits, our gates and
circuits are intentionally stochastic: each output is a sample from a
probability distribution conditioned on the inputs, and (except in
degenerate cases) simulating a circuit twice will produce different
results. The numerical probability distributions themselves are
implicit, though they can be estimated via the circuits' long-run
time-averaged behavior. And also unlike digital gates and circuits,
Bayesian reasoning arises naturally via the dynamics of our
synchronously clocked circuits, simply by fixing the values of the
circuit elements representing the data.

We have built prototype circuits that solve problems of depth and
motion perception and perceptual learning, plus a compiler that can
automatically generate circuits for solving causal reasoning problems
given a description of the underlying causal model. Each of these
systems illustrates the use of stochastic digital circuits to
accelerate Bayesian inference an important class of probabilistic
models, including Markov Random Fields, nonparametric Bayesian mixture
models, and Bayesian networks. Our prototypes show that this
combination of simple choices at the hardware level --- a discrete,
digital representation for information, coupled with intentionally
stochastic rather than ideally deterministic elements --- has far
reaching architectural consequences. For example, software
implementations of approximate Bayesian reasoning typically rely on
high-precision arithmetic and serial computation. We show that our
synchronous stochastic circuits can be implemented at very low bit
precision, incurring only a negligible decrease in accuracy. This low
precision enables us to make fast, small, power-efficient circuits at
the core of our designs. We also show that these reductions in
computing unit size are sufficient to let us exploit the massive
parallelism that has always been inherent in complex probabilistic
models at a granularity that has been previously impossible to
exploit. The resulting high computation density drives the performance
gains we see from stochastic digital circuits, narrowing the
efficiency gap with neural computation by multiple orders of
magnitude.

Our approach is fundamentally different from existing approaches for
reliable computation with unreliable
components\cite{neumanncomputer,akgulprobabilistic,gaines1969stochastic},
which view randomness as either a source of error whose impact needs
to be mitigated or as a mechanism for approximating arithmetic
calculations. Our combinational circuits are intentionally stochastic,
and we depend on them to produce exact samples from the probability
distributions they represent. Our approach is also different from and
complementary to classic analog\cite{mead1990neuromorphic} and modern
mixed-signal\cite{choudhary2012silicon} neuromorphic computing
approaches: stochastic digital primitives and architectures could
potentially be implemented using neuromorphic techniques, providing a
means of applying these designs to problems of Bayesian inference.

In theory, stochastic digital circuits could be used to solve any
computable Bayesian inference problem with a computable
likelihood\cite{afrcomputable} by implementing a Markov chain for
inference in a Turing-complete probabilistic programming
language\cite{mansinghka2009natively, Goodman:2008tb}. Stochastic
ciruits can thus implement inference and learning techniques for
diverse intelligent computing architectures, including both
probabilistic models defined over structured, symbolic
representations\cite{tenenbaumgrow} as well as sparse, distributed,
connectionist representations\cite{SalHinton07}. In contrast, hardware
accelerators for belief propagation
algorithms\cite{pearlprobabilistic, linhigh, vigodacontinuous} can
only answer queries about marginal probabilities or most probable
configurations, only apply to finite graphical models with discrete or
binary nodes, and cannot be used to learn model parameters from
data. For example, the formulation of perceptual learning we present
here is based on inference in a nonparametric Bayesian model to which
belief propagation does not apply. Additionally, because stochastic
digital circuits produce samples rather than probabilities, their
results capture the complex dependencies between variables in
multi-modal probability distributions, and can also be used to solve
otherwise intractable problems in decision theory by estimating
expected utilities.

\section*{Stochastic Digital Gates and Stateless Stochastic Circuits}

%\begin{figure*}[p]
%\begin{center}
{\bf (Figure 1 about here)}
\label{fig:combinational}
%
%\end{center}
%\end{figure*}

Digital logic circuits are based on a gate abstraction defined by
Boolean functions: deterministic mappings from input bit values to
output bit values\cite{shannonsymbolic}. For elementary gates, such
as the AND gate, these are given by truth tables; see Figure 1A. Their
power and flexibility comes in part from the composition laws that
they support, shown in Figure 1B. The output from one gate can be
connected to the input of another, yielding a circuit that samples
from the composition of the Boolean functions represented by each
gate. The compound circuit can also be treated as a new primitive,
abstracting away its internal structure. These simple laws have proved
surprisingly powerful: they enable complex circuits to be built up out
of reusable pieces.

{\em Stochastic digital gates} (see Figure 1C) are similar to Boolean
gates, but consume a source of random bits to generate samples from
conditional probability distributions. Stochastic gates are specified
by conditional probability tables; these give the probability that a
given output will result from a given input. Digital logic corresponds
to the degenerate case where all the probabilities are 0 or 1; see
Figure 1D for the conditional probability table for an AND gate. Many
stochastic gates with m input bits and n output bits are
possible. Figure 1E shows one central example, the THETA gate, which
generates draws from a biased coin whose bias is specified on the
% SUPPLEMENTARY
input. Supplementary material outlining serial and parallel
implementations is available at
\cite{VMEJ-circuits-supplemental}. Crucially, stochastic gates support
generalizations of the composition laws from digital logic, shown in
Figure 1F. The output of one stochastic gate can be fed as the input
to another, yielding samples from the joint probability distribution
over the random variables simulated by each gate. The compound circuit
can also be treated as a new primitive that generates samples from the
marginal distribution of the final output given the first input. As
with digital gates, an enormous variety of circuits can be constructed
using these simple rules.

\section*{Fast Bayesian Inference via Massively Parallel Stochastic Transition Circuits}

Most digital systems are based on deterministic finite state machines;
the template for these machines is shown in Figure 2A. A stateless
digital circuit encodes the transition function that calculates the
next state from the previous state, and the clocking machinery (not
shown) iterates the transition function repeatedly. This abstraction
has proved enormously fruitful; the first microprocessors had roughly
$2^{20}$ distinct states. In Figure 2B, we show the stochastic analogue
of this synchronous state machine: a {\em stochastic transition
  circuit}.

Instead of the combinational logic circuit implementing a
deterministic transition function, it contains a combinational
stochastic circuit implementing a stochastic transition operator that
samples the next state from a probability distribution that depends on
the current state. It thus corresponds to a Markov chain in
hardware. To be a valid transition circuit, this transition operator
must have a unique stationary distribution $P(S|X)$ to which it
ergodically converges. A number of recipes for suitable transition
operators can be constructed, such as Metropolis sampling
\cite{metropolisequation} and Gibbs sampling\cite{gemanstochastic};
most of the results we present rely on variations on Gibbs
% SUPPLEMENTARY
sampling. More details on efficient implementations of stochastic
transition circuits for Gibbs sampling and Metropolis-Hastings can be
found elsewhere \cite{VMEJ-circuits-supplemental}. Note that if the
input $X$ represents observed data and the state $S$ represents a
hypothesis, then the transition circuit implements Bayesian inference.

We can scale up to challenging problems by exploiting the composition
laws that stochastic transition circuits support. Consider a
probability distribution defined over three variables $P(A,B,C) =
P(A)P(B|A)P(C|A)$. We can construct a transition circuit that samples
from the overall state $(A,B,C)$ by composing transition circuits for
updating $A|BC$, $B|A$ and $C|A$; this assembly is shown in Figure
2C. As long as the underlying probability model does not have any
zero-probability states, ergodic convergence of each constituent
transition circuit then implies ergodic convergence of the whole
assembly\cite{andrieuintroduction}. The only requirement for
scheduling transitions is that each circuit must be left fixed while
circuits for variables that interact with it are transitioning. This
scheduling requirement --- that a transition circuit's value be held
fixed while others that read from its internal state or serve as
inputs to its next transition are updating --- is analogous to the
so-called ``dynamic discipline'' that defines valid clock schedules
for traditional sequential
logic\cite{ward1990computation}. Deterministic and stochastic
schedules, implementing cycle or mixture hybrid
kernels\cite{andrieuintroduction}, are both possible. This simple rule
also implies a tremendous amount of exploitable parallelism in
stochastic transition circuits: if two variables are independently
caused given the current setting of all others, they can be updated at
the same time.

Assemblies of stochastic transition circuits implement Bayesian
reasoning in a straightforward way: by fixing, or ``clamping'' some of
the variables in the assembly. If no variables are fixed, the circuit
explores the full joint distribution, as shown in Figure 2E and 2F. If
a variable is fixed, the circuit explores the conditional distribution
on the remaining variables, as shown in Figure 2G and 2H. Simply by
changing which transition circuits are updated, the circuit can be
used to answer different probabilistic queries; these can be varied
online based on the needs of the application.

%\begin{figure*}
%\begin{center}
{\bf (Figure 2 about here.)}
\label{fig:transition}
%\end{center}
%\end{figure*}

\section*{The accuracy of ultra-low-precision stochastic transition circuits.}

The central operation in many Markov chain techniques for inference is
called DISCRETE-SAMPLE, which generates draws from a discrete-output
probability distribution whose weights are specified on its input. For
example, in Gibbs sampling, this distribution is the conditional
probability of one variable given the current value of all other
variables that directly depend on it. One implementation of this
operation is shown in Figure 3A; each stochastic transition circuit
from Figure 2 could be implemented by one such circuit, with
multiplexers to select log-probability values based on the neighbors
of each random variable. Because only the ratios of the raw
probabilities matter, and the probabilities themselves naturally vary
on a log scale, extremely low precision representations can still
provide accurate results. High entropy (i.e. nearly uniform)
distributions are resilient to truncation because their values are
nearly equal to begin with, differing only slightly in terms of their
low-order bits. Low entropy (i.e. nearly deterministic) distributions
are resilient because truncation is unlikely to change which outcomes
have nonzero probability. Figure 3B quantifies this low-precision
property, showing the relative entropy (a canonical information
theoretic measure of the difference between two distributions) between
the output distributions of low precision implementations of the
circuit from Figure 3A and an accurate floating-point
implementation. Discrete distributions on 1000 outcomes were used,
spanning the full range of possible entropies, from almost 10 bits
(for a uniform distribution on 1000 outcomes) to 0 bits (for a
deterministic distribution), with error nearly undetectable until
fewer than 8 bits are used. Figure 3C shows example distributions on
10 outcomes, and Figure 3D shows the resulting impact on computing
element size. Extensive quantitative assessments of the impact of low
bit precision have also been performed, providing additional evidence
that only very low precision is required
\cite{VMEJ-circuits-supplemental}.
% SUPPLEMENTARY

%\begin{figure*}
%\begin{center}
{\bf (Figure 3 about here.)}
\label{fig:MULTINOMIAL}
%\end{center}
%\end{figure*}

\section*{Efficiency gains on depth and motion perception and perceptual learning problems}

Our main results are based on an implementation where each stochastic
gate is simulated using digital logic, consuming entropy from an
internal pseudorandom number generator\cite{marsagliaxorshift}. This
allows us to measure the performance and fault-tolerance improvements
that flow from stochastic architectures, independent of physical
implementation. We find that stochastic circuits make it practical to
perform stochastic inference over several probabilistic models with
10,000+ latent variables in real time and at low power on a single
chip. These designs achieve a 1,000x speed advantage over commodity
microprocessors, despite using gates that are 10x slower.
% WISHLIST Xilinx citation that works\cite{xilinxds}
% The above citation is of the dc and swithcing characteristics of the virtex-6 FPGA that we're using, which contains information about maximum clock rate. 
% SUPPLEMENTARY
In \cite{VMEJ-circuits-supplemental}, we also show architectures that
exhibit minimal degradation of accuracy in the presence of fault rates
as high as one bit error for every 100 state transitions, in contrast
to conventional architectures where failure rates are measured in bit
errors (failures) per billion hours of operation\cite{bertrillions}.

Our first application is to depth and motion perception, via Bayesian
inference in lattice Markov Random Field models\cite{gemanstochastic}. The core problem is matching pixels from two
images of the same scene, taken at distinct but nearby points in space
or in time. The matching is ambiguous on the basis of the images
alone, as multiple pixels might share the same value\cite{marrcooperative}; prior knowledge about the structure of the
scene must be applied, which is often cast in terms of Bayesian
inference\cite{szeliskicomparative}. Figure 4A illustrates the
  template probabilistic model most commonly used. The X variables
  contain the unknown displacement vectors. Each Y variable contains a
  vector of pixel similarity measurements, one per possible pair of
  matched pixels based on X. The pairwise potentials between the X
  variables encode scene structure assumptions; in typical problems,
  unknown values are assumed to vary smoothly across the scene, with a
  small number of discontinuities at the boundaries of objects. Figure
  4B shows the conditional independence structure in this problem:
  every other X variable is independent from one another, allowing the
  entire Markov chain over the X variables to be updated in a
  two-phase clock, independent of lattice size. Figure 4C shows the
  dataflow for the software-reprogrammable probabilistic video
  processor we developed to solve this family of problems; this
  processor takes a problem specification based on pairwise potentials
  and Y values, and produces a stream of posterior samples. When
  comparing the hardware to hand-optimized C versions on a commodity
  workstation, we see a 500x performance improvement.

%\begin{figure*}
%\begin{center}
%\vspace{-0.5in}
{\bf (Figure 4 about here.)}
%\vspace{-1in}
\label{fig:VISION}
%\end{center}
%\end{figure*}

We have also built stochastic architectures for solving perceptual
learning problems, based on fully Bayesian inference in Dirichlet
process mixture
models\cite{fergusonbayesian,rasmusseninfinite}. Dirichlet process
mixtures allow the number of clusters in a perceptual dataset to be
automatically discovered during inference, without assuming an a
priori limit on the models' complexity, and form the basis of many
models of human
categorization\cite{andersonrational,griffithscategorization}. We
tested our prototype on the problem of discovering and classifying
handwritten digits from binary input images. Our circuit for solving
this problem operates on an online data stream, and efficiently tracks
the number of perceptual clusters this input; see
% SUPPLEMENTARY
\cite{VMEJ-circuits-supplemental} for architectural and implementation
details and additional characterizations of performance. As with our
depth and motion perception architecture, we observe over $\sim$2,000x
speedups as compared to a highly optimized software implementation. Of
the $\sim$2000x difference in speed, roughly $\sim$256x is directly
due to parallelism --- all of the pixels are independent dimensions,
and can therefore be updated simultaneously.

%\begin{figure*}
%\begin{center}
{\bf (Figure 5 about here.)}
\label{fig:learning}
%\end{center}
%\end{figure*}

\section*{Automatically generated causal reasoning circuits and spiking implementations}

Digital logic gates and their associated design rules are so simple
that circuits for many problems can be generated
automatically. Digital logic also provides a common target for device
engineers, and have been implemented using many different physical
mechanisms -- classically with vaccum tubes, then with MOSFETS in
silicon, and even on spintronic devices\cite{spintronicnand}. Here we
provide two illustrations of the analogous simplicity and generality
of stochastic digital circuits, both relevant for the
reverse-engineering of intelligent computation in the brain.

We have built a compiler that can automatically generate circuits for
solving arbitrary causal reasoning problems in Bayesian network
models. Bayesian network formulations of causal reasoning have played
central roles in machine intelligence\cite{pearlprobabilistic} and
computational models of cognition in both humans and rodents\cite{blaisdellcausal}. Figure \ref{fig:COMPILER}A shows a Bayesian
network for diagnosing the behavior of an intensive care unit
monitoring system. Bayesian inference within this network can be used
to infer probable states of the ICU given ambiguous patterns of
evidence --- that is, reason from observed effects back to their
probable causes. Figure \ref{fig:COMPILER}B shows a factor graph
representation of this model\cite{kschischangfactor}; this more
general data structure is used as the input to our compiler. Figure
\ref{fig:COMPILER}C shows inference results from three representative
queries, each corresponding to a different pattern of observed data.

We have also explored implementations of stochastic transition
circuits in terms of spiking elements governed by Poisson firing
statistics. Figure \ref{fig:COMPILER}D shows a spiking network that
implements the Markov chain from Figure \ref{fig:transition}. The
stochastic transition circuit corresopnding to a latent variable $X$
is implemented via a bank of Poisson-spiking elements $\{X_i\}$ with
one unit $X_i$ per possible value of the variable. The rate for each
spiking element $X_i$ is determined by the unnormalized conditional
log probability of the variable setting it corresponds to, following
the discrete-sample gate from Figure~\ref{fig:MULTINOMIAL} the time to
first spike $\mathrm{t}(X_i) \sim Exp(e_i)$, with $e_i$ obtained by
summing energy contributions from all connected variables. The output
value of $X$ is determined by $\mathrm{argmin}_i \{\mathrm{t}(X_i)\}$,
i.e. the element that spiked first, implemented by fast lateral
inhibition between the $X_i$s. It is easy to show that this implements
% SUPPLEMENT
exponentiation and normalization of the energies, leading to a correct
implementation of a stochastic transition circuit for Gibbs sampling;
see \cite{VMEJ-circuits-supplemental} for more information. Elements are
clocked quasi-synchronously, reflecting the conditional independence
structure and parallel update scheme from
Figure~\ref{fig:transition}D, and yields samples from the correct
equilibrium distribution.

This spiking implementation helps to narrow the gap with recent
theories in computational neuroscience. For example, there have been
recent proposals that neural spikes correspond to
samples\cite{fiser2010statistically}, and that some spontaneous
spiking activity corresponds to sampling from the brain's unclamped
prior distribution\cite{berkes2011spontaneous}. Combining these local
elements using our composition and abstraction laws into massively
parallel, low-precision, intentionally stochastic circuits may help to
bridge the gap between probabilistic theories of neural computation
and the computational demands of complex probabilistic models and
approximate inference\cite{probbrains}.
%A spike-time representation
%has greater information carrying capacity than a rate code
%\cite{van2001rate}. 
 
%\begin{figure*}
%\begin{center}
{\bf (Figure 6 about here.)}
\label{fig:COMPILER}
%\end{center}
%\end{figure*}

\section*{Discussion}

To further narrow the efficiency gap with the brain, and scale to more
challenging Bayesian inference problems, we need to improve the
convergence rate of our architectures. One approach would be to
initialize the state in a transition circuit via a separate,
feed-forward, combinational circuit that approximates the equilibrium
distribution of the Markov chain. Machine perception software that
uses machine learning to construct fast, compact initializers is
already in use\cite{shottonreal}. Analyzing the number of transitions
needed to close the gap between a good initialization and the target
distribution may be harder\cite{diaconismarkov}. However, some
feedforward Monte Carlo inference strategies for Bayesian networks
provably yield precise estimates of probabilities in polynomial time
if the underlying probability model is sufficiently stochastic\cite{dagum1997optimal}; it remains to be seen if similar conditions
apply to stateful stochastic transition circuits.

It may also be fruitful to search for novel electronic devices --- or
previously unusable dynamical regimes of existing devices --- that are
as well matched to the needs of intentionally stochastic circuits as
transistors are to logical inverters, potentially even via a spiking
implementation. Physical phenomena that proved too unreliable for
implementing Boolean logic gates may be viable building blocks for
machines that perform Bayesian inference.

Computer engineering has thus far focused on deterministic mechanisms
of remarkable scale and complexity: billlions of parts that are
expected to make trillions of state transitions with perfect
repeatability\cite{intelInstructionErrors}. But we are now
engineering computing systems to exhibit more intelligence than they
once did, and identify probable explanations for noisy, ambiguous
data, drawn from large spaces of possibilities, rather than calculate
the definite consequences of perfectly known assumptions with high
precision. The apparent intractability of probabilistic inference has
complicated these efforts, and challenged the viability of Bayesian
reasoning as a foundation for engineering intelligent computation and
for reverse-engineering the mind and brain.

At the same time, maintaining the illusion of rock-solid determinism
has become increasingly costly. Engineers now attempt to build digital
logic circuits in the deep sub-micron regime\cite{shepardnoise} and
even inside cells\cite{elowitzsynthetic}; in both these settings, the
underlying physics has stochasticity that is difficult to
suppress. Energy budgets have grown increasingly restricted, from the
scale of the datacenter\cite{barroso2007case} to the mobile
device\cite{flinn1999energy}, yet we spend substantial energy to
operate transistors in deterministic regimes. And efforts to
understand the dynamics of biological computation --- from biological
neural networks to gene expression
networks\cite{mcadams1997stochastic} --- have all encountered
stochastic behavior that is hard to explain in deterministic, digital
terms. Our intentionally stochastic digital circuit elements and
stochastic computing architectures suggest a new direction for
reconciling these trends, and enables the design of a new class of
fast, Bayesian digital computing machines.

% [[Wishlist: cite markram2006blue, de2010world]]
% Wishlist: themes of connectionism, leaving aside the physical to focus on the abstract computational implications

\begin{addendum}
 \item[Acknowledgements] The authors would like to acknowledge Tomaso
   Poggio, Thomas Knight, Gerald Sussman, Rakesh Kumar and Joshua
   Tenenbaum for numerous helpful discussions and comments on early
   drafts, and Tejas Kulkarni for contributions to the spiking
   implementation.
\end{addendum}

\pagebreak
\noindent {\bf Figure 1.} {\small (A) Boolean gates, such as the AND gate, are mathematically specified by truth tables: deterministic mappings from binary inputs to binary outputs. (B) Compound Boolean circuits can be synthesized out of sub-circuits that  each calculate different sub-functions, and treated as a single gate that implements the composite function, without reference to its internal details. (C) Each stochastic gate samples from a discrete probability distribution conditioned on an input; for clarity, we show an external source of random bits driving the stochastic behavior. (D) Composing gates that sample B given A and C given B yields a network that samples from the joint distribution over B and C given A; abstraction yields a gate that samples from the marginal distribution C|A. When only one sample path has nonzero probability, this recovers the composition of Boolean functions. (E) The THETA gate is a stochastic gate that generates samples from a Bernoulli distribution whose parameter theta is specified via the $m$ input bits. Like all stochastic digital gates, it can be specified by a conditional probability table, analogously to how Boolean gates can be specified via a truth table. (F) When each new output sample is triggered (e.g. because its internal randomness source updates), a different output sample is generated; time-averaging the output makes it possible to estimate the entries in the probability table, which are otherwise implicit. (G) The THETA gate can be implemented by comparing the output of a source of (pseudo)random bits to the input coin weight. (H) Deterministic gates, such as the AND gate shown here, can be viewed as degenerate stochastic gates specified by conditional probability tables whose entries are either 0 or 1. This permits fluid interoperation of deterministic and stochastic gates in compound circuits. (I) A parallel circuit implementing a Binomial random variable can be implemented by combining THETA gates and adders using the composition laws from (D).}

\pagebreak
\includegraphics[width=\textwidth]{figures/f1-combinational.ai}

\pagebreak
\noindent {\bf Figure 2.} {\small Stochastic transition circuits and massively parallel Bayesian inference. (A) A deterministic finite state machine consists of a register and a transition function implemented via combinational logic. (B) A stochastic transition circuit consists of a register and a stochastic transition operator implemented by a combinational stochastic circuit. Each stochastic transition circuit is $T_{S|X}$ is parameterized by some input $X$, and its internal combinational stochastic block $P(S_{t+1}|S_t,X)$ must ergodically converge to a unique stationary distribution $P(S|X)$ for all $X$. (C) Stochastic transition circuits can be composed to construct samplers for probabilistic models over multiple variables by wiring together stochastic transition circuits for each variable based on their interactions. This circuit samples from a distribution $P(A,B,C) = P(A)P(B|A)P(C|A)$. (D) Each network of stochastic transition circuits can be scheduled in many ways; here we show one serial schedule and one parallel schedule for the transition circuit from (C). Convergence depends only on respecting the invariant that no stochastic transition circuit transitions while other circuits that interact with it are transitioning. (E) The Markov chain implemented by this transition circuit. (F) Typical stochastic evolutions of the state in this circuit. (G) Inference can be implemented by clamping state variables to specific values; this yields a restricted Markov chain that converges to the conditional distribution over the unclamped variables given the clamped ones. Here we show the chain obtained by fixing $C=1$. (H) Typical stochastic evolutions of the state in this clamped transition circuit. Changing which variables are fixed allows the inference problem to be changed dynamically as the circuit is running.}

\pagebreak
\includegraphics[height=\textheight]{figures/f2-transitions.ai}

\pagebreak
\noindent {\small {\bf Figure 3.} (A) The discrete-sample gate is a central building
  block for stochastic transition circuits, used to implement Gibbs
  transition operators that update a variable by sampling from its
  conditional distribution given the variables it interacts with. The
  gate renormalizes the input log probabilities it is given, converts
  them to probabilities (by exponentiation), and then samples from the
  resulting distribution. Input energies are specified via a custom
  fixed-point coding scheme. (B) Discrete-sample gates remain accurate
  even when implemented at extremely low bit-precision. Here we show
  the relative entropy between true distributions and their
  low-precision implementations, for millions of distributions over
  discrete sets with 1000 elements; accuracy loss is negligible even
  when only 8 bits of precision are used. (C) The accuracy of
  low-precision discrete-sample gates can be understood by considering
  multinomial distributions with high, medium and low entropy. High
  entropy distributions involve outcomes with very similar
  probability, insensitive to ratios, while low entropy distributions
  are dominated by the location of the most probable outcome. (D)
  Low-precision transition circuits save area as compared to
  high-precision floating point alternatives; these area savings make
  it possible to economically exploit massive parallelism, by fitting
  many sampling units on a single chip.}

\pagebreak
\includegraphics[width=\textwidth]{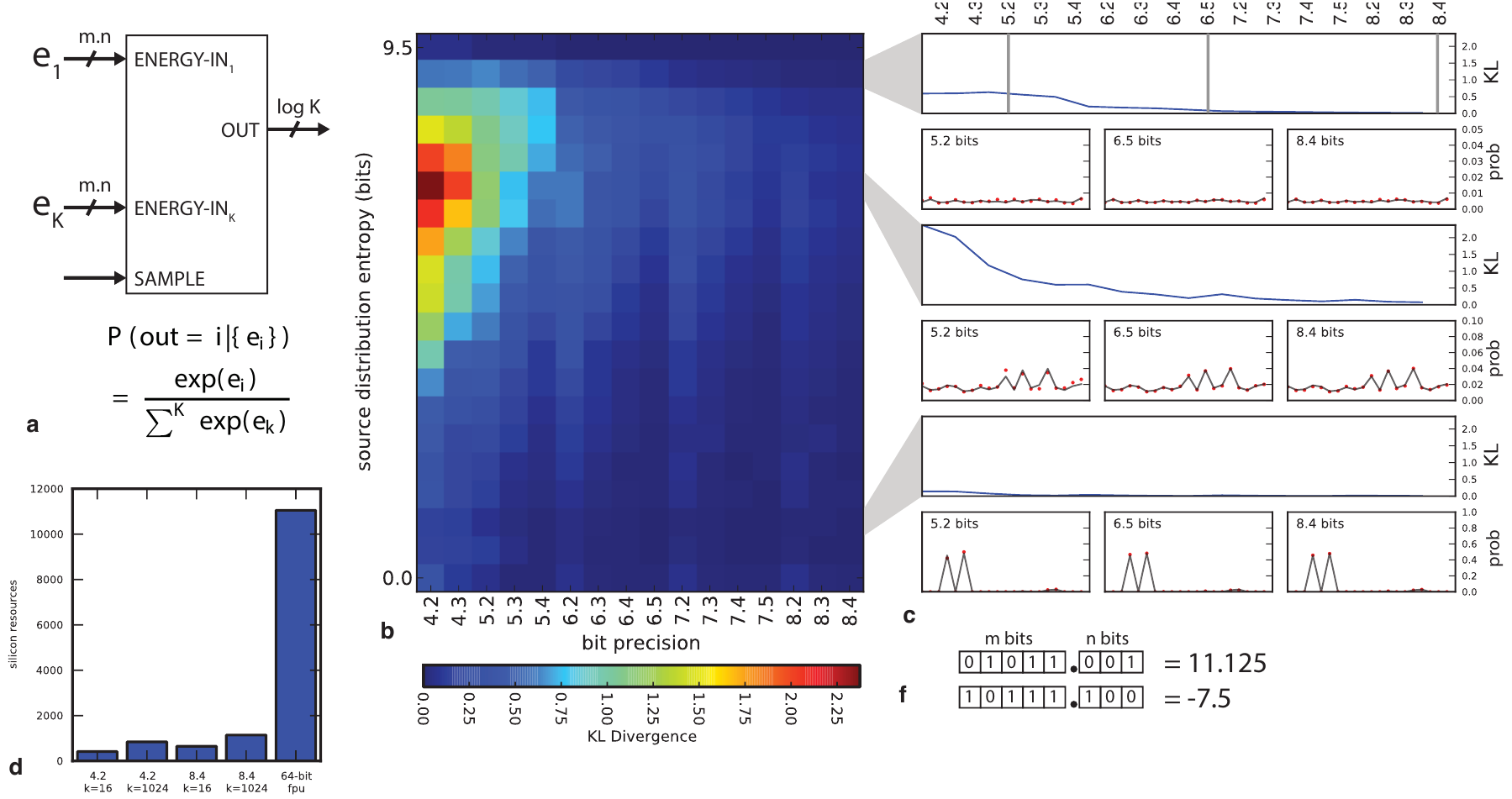}

\pagebreak
\noindent {\small {\bf Figure 4.} (A) A Markov Random Field for
  solving depth and motion perception, as well as other dense matching
  problems. Each $X_{i,j}$ node stores the hidden quantity to be
  estimated, e.g. the disparity of a pixel. Each $f_{LP}$ ensures
  adjacent $X$s are either similar or very different, i.e. that depth
  and motion fields vary smoothly on objects but can contain
  discontinuities at object boundaries. Each $Y_{i,j}$ node stores a
  per-latent-pixel vector of similarity information for a range of
  candidate matches, linked to the $X$s by the $f_E$ potentials. (B)
  The conditional independencies in this model permit many different
  parallelization strategies, from fully space-parallel
  implementations to virtualized implementations where blocks of
  pixels are updated in parallel. (C) Depth perception results. The
  left input image, plus the depth maps obtained by software (middle)
  and hardware (right) engines for solving the Markov Random
  Field. (D) Motion perception results. One input frame, plus the
  motion flow vector fields for software (middle) and hardware (right)
  solutions. (E) Energy versus time for software and hardware
  solutions to depth perception, including both 8-bit and 12-bit
  hardware. Note that the hardware is roughly 500x faster than the
  software on this frame. (F) Energy versus time for software and
  hardware solutions to motion perception.}

\pagebreak
\includegraphics[width=\textwidth]{figures/f4-vision.ai}

\pagebreak
\noindent {\bf Figure 5.} {\small (A) Example samples from the posterior distribution
  of cluster assignments for a nonparametric mixture model. The two samples
  show posterior variance, reflecting the uncertainty between three
  and four source clusters. 
  (B) Typical handwritten digit images
  from the MNIST corpus\cite{lecun1998mnist}, showing a high degree of
  variation across digits of the same type. (C) The digit clusters
  discovered automatically by a stochastic digital circuit for
  inference in Dirichlet process mixture models. Each image represents
  a cluster; each pixel represents the probability that the
  corresponding image pixel is black. Clusters are sorted according to
  the most probable true digit label of the images in the
  cluster. Note that these cluster labels were not provided to the
  circuit. Both the clusters and the number of clusters were
  discovered automatically by the circuit over the course of
  inference. (D) The receiver operating characteristic (ROC) curves
  that result from classifying digits using the learned clusters;
  quantitative results are competitive with state-of-the-art
  classifiers. (E) The time required for one cycle through the
  outermost transition circuit in hardware, versus the corresponding
  time for one sweep of a highly optimized software implementation of
  the same sampler, which is $\sim$2000x slower.}

\pagebreak
\includegraphics[width=\textwidth]{figures/f5-learning.ai}

\pagebreak
\noindent {\bf Figure 6.} {\small (A) A Bayesian network model for ICU alarm monitoring, showing measurable variables, hidden variables, and diagnostic variables of interest. (B) A factor graph representation of this Bayesian network, rendered by the input stage for our stochastic transition circuit synthesis software. (C) A representation of the factor graph showing evidence variables as well as a parallel schedule for the transition circuits automatically extracted by our compiler: all nodes of the same color can be transitioned simultaneously. (D) Three diagnosis results from Bayesian inference in the alarm network, showing high accuracy diagnoses (with some posterior uncertainty) from an automatically generated circuit. E) The schematic of a spiking neural implementation of a stochastic transition circuit assembly for sampling from the three-variable probabilistic model from Figure 2. (F) The spike raster (black) and state sequence (blue) that result from simulating the circuit. (G) The spiking simulation yields state distributions that agree with exact simulation of the underlying Markov chain.}

\pagebreak
\includegraphics[width=\textwidth]{figures/f6-compiler.ai}

\bibliography{VMEJ-circuits-arxiv}

\begin{thebibliography}{10}
\expandafter\ifx\csname url\endcsname\relax
  \def\url#1{\texttt{#1}}\fi
\expandafter\ifx\csname urlprefix\endcsname\relax\def\urlprefix{URL }\fi
\providecommand{\bibinfo}[2]{#2}
\providecommand{\eprint}[2][]{\url{#2}}

\bibitem{weissmotion}
\bibinfo{author}{Weiss, Y.}, \bibinfo{author}{Simoncelli, E.~P.} \&
  \bibinfo{author}{Adelson, E.~H.}
\newblock \bibinfo{title}{Motion illusions as optimal percepts}.
\newblock \emph{\bibinfo{journal}{Nature neuroscience}}
  \textbf{\bibinfo{volume}{5}}, \bibinfo{pages}{598--604}
  (\bibinfo{year}{2002}).

\bibitem{kordinbayesian}
\bibinfo{author}{K{\"o}rding, K.~P.} \& \bibinfo{author}{Wolpert, D.~M.}
\newblock \bibinfo{title}{Bayesian integration in sensorimotor learning}.
\newblock \emph{\bibinfo{journal}{Nature}} \textbf{\bibinfo{volume}{427}},
  \bibinfo{pages}{244--247} (\bibinfo{year}{2004}).

\bibitem{griffithsoptimal}
\bibinfo{author}{Griffiths, T.~L.} \& \bibinfo{author}{Tenenbaum, J.~B.}
\newblock \bibinfo{title}{Optimal predictions in everyday cognition}.
\newblock \emph{\bibinfo{journal}{Psychological Science}}
  \textbf{\bibinfo{volume}{17}}, \bibinfo{pages}{767--773}
  (\bibinfo{year}{2006}).

\bibitem{blaisdellcausal}
\bibinfo{author}{Blaisdell, A.~P.}, \bibinfo{author}{Sawa, K.},
  \bibinfo{author}{Leising, K.~J.} \& \bibinfo{author}{Waldmann, M.~R.}
\newblock \bibinfo{title}{Causal reasoning in rats}.
\newblock \emph{\bibinfo{journal}{Science}} \textbf{\bibinfo{volume}{311}},
  \bibinfo{pages}{1020--1022} (\bibinfo{year}{2006}).

\bibitem{tenenbaumgrow}
\bibinfo{author}{Tenenbaum, J.~B.}, \bibinfo{author}{Kemp, C.},
  \bibinfo{author}{Griffiths, T.~L.} \& \bibinfo{author}{Goodman, N.~D.}
\newblock \bibinfo{title}{How to grow a mind: Statistics, structure, and
  abstraction}.
\newblock \emph{\bibinfo{journal}{science}} \textbf{\bibinfo{volume}{331}},
  \bibinfo{pages}{1279--1285} (\bibinfo{year}{2011}).

\bibitem{ferruccibuilding}
\bibinfo{author}{Ferrucci, D.} \emph{et~al.}
\newblock \bibinfo{title}{Building watson: An overview of the deepqa project}.
\newblock \emph{\bibinfo{journal}{AI magazine}} \textbf{\bibinfo{volume}{31}},
  \bibinfo{pages}{59--79} (\bibinfo{year}{2010}).

\bibitem{thrunprobabilisticarticle}
\bibinfo{author}{Thrun, S.}
\newblock \bibinfo{title}{Probabilistic robotics}.
\newblock \emph{\bibinfo{journal}{Communications of the ACM}}
  \textbf{\bibinfo{volume}{45}}, \bibinfo{pages}{52--57}
  (\bibinfo{year}{2002}).

\bibitem{thrunprobabilisticbook}
\bibinfo{author}{Thrun, S.}, \bibinfo{author}{Burgard, W.},
  \bibinfo{author}{Fox, D.} \emph{et~al.}
\newblock \emph{\bibinfo{title}{Probabilistic robotics}},
  vol.~\bibinfo{volume}{1} (\bibinfo{publisher}{MIT press Cambridge},
  \bibinfo{year}{2005}).

\bibitem{shottonreal}
\bibinfo{author}{Shotton, J.} \emph{et~al.}
\newblock \bibinfo{title}{Real-time human pose recognition in parts from single
  depth images}.
\newblock \emph{\bibinfo{journal}{Communications of the ACM}}
  \textbf{\bibinfo{volume}{56}}, \bibinfo{pages}{116--124}
  (\bibinfo{year}{2013}).

\bibitem{univac}
\bibinfo{author}{Eckert~Jr, J.~P.}, \bibinfo{author}{Weiner, J.~R.},
  \bibinfo{author}{Welsh, H.~F.} \& \bibinfo{author}{Mitchell, H.~F.}
\newblock \bibinfo{title}{The univac system}.
\newblock In \emph{\bibinfo{booktitle}{Papers and discussions presented at the
  Dec. 10-12, 1951, joint AIEE-IRE computer conference: Review of electronic
  digital computers}}, \bibinfo{pages}{6--16} (\bibinfo{organization}{ACM},
  \bibinfo{year}{1951}).

\bibitem{shivakumar2002modeling}
\bibinfo{author}{Shivakumar, P.}, \bibinfo{author}{Kistler, M.},
  \bibinfo{author}{Keckler, S.~W.}, \bibinfo{author}{Burger, D.} \&
  \bibinfo{author}{Alvisi, L.}
\newblock \bibinfo{title}{Modeling the effect of technology trends on the soft
  error rate of combinational logic}.
\newblock In \emph{\bibinfo{booktitle}{Dependable Systems and Networks, 2002.
  DSN 2002. Proceedings. International Conference on}},
  \bibinfo{pages}{389--398} (\bibinfo{organization}{IEEE},
  \bibinfo{year}{2002}).

\bibitem{synapticfailure}
\bibinfo{author}{Rosenmund, C.}, \bibinfo{author}{Clements, J.} \&
  \bibinfo{author}{Westbrook, G.}
\newblock \bibinfo{title}{Nonuniform probability of glutamate release at a
  hippocampal synapse}.
\newblock \emph{\bibinfo{journal}{Science}} \textbf{\bibinfo{volume}{262}},
  \bibinfo{pages}{754--757} (\bibinfo{year}{1993}).

\bibitem{neumanncomputer}
\bibinfo{author}{Neumann, J.~v.}
\newblock \bibinfo{title}{The computer and the brain}  (\bibinfo{year}{1958}).

\bibitem{akgulprobabilistic}
\bibinfo{author}{Akgul, B.~E.}, \bibinfo{author}{Chakrapani, L.~N.},
  \bibinfo{author}{Korkmaz, P.} \& \bibinfo{author}{Palem, K.~V.}
\newblock \bibinfo{title}{Probabilistic cmos technology: A survey and future
  directions}.
\newblock In \emph{\bibinfo{booktitle}{Very Large Scale Integration, 2006 IFIP
  International Conference on}}, \bibinfo{pages}{1--6}
  (\bibinfo{organization}{IEEE}, \bibinfo{year}{2006}).

\bibitem{gaines1969stochastic}
\bibinfo{author}{Gaines, B.}
\newblock \bibinfo{title}{Stochastic computing systems}.
\newblock \emph{\bibinfo{journal}{Advances in information systems science}}
  \textbf{\bibinfo{volume}{2}}, \bibinfo{pages}{37--172}
  (\bibinfo{year}{1969}).

\bibitem{mead1990neuromorphic}
\bibinfo{author}{Mead, C.}
\newblock \bibinfo{title}{Neuromorphic electronic systems}.
\newblock \emph{\bibinfo{journal}{Proceedings of the IEEE}}
  \textbf{\bibinfo{volume}{78}}, \bibinfo{pages}{1629--1636}
  (\bibinfo{year}{1990}).

\bibitem{choudhary2012silicon}
\bibinfo{author}{Choudhary, S.} \emph{et~al.}
\newblock \bibinfo{title}{Silicon neurons that compute}.
\newblock In \emph{\bibinfo{booktitle}{Artificial Neural Networks and Machine
  Learning--ICANN 2012}}, \bibinfo{pages}{121--128}
  (\bibinfo{publisher}{Springer}, \bibinfo{year}{2012}).

\bibitem{afrcomputable}
\bibinfo{author}{{Ackerman}, N.~L.}, \bibinfo{author}{{Freer}, C.~E.} \&
  \bibinfo{author}{{Roy}, D.~M.}
\newblock \bibinfo{title}{{On the computability of conditional probability}}.
\newblock \emph{\bibinfo{journal}{ArXiv e-prints}}  (\bibinfo{year}{2010}).
\newblock \eprint{1005.3014}.

\bibitem{mansinghka2009natively}
\bibinfo{author}{Mansinghka, V.~K.}
\newblock \emph{\bibinfo{title}{Natively probabilistic computation}}.
\newblock Ph.D. thesis, \bibinfo{school}{Massachusetts Institute of Technology}
  (\bibinfo{year}{2009}).

\bibitem{Goodman:2008tb}
\bibinfo{author}{Goodman, N.~D.}, \bibinfo{author}{Mansinghka, V.~K.},
  \bibinfo{author}{Roy, D.~M.}, \bibinfo{author}{Bonowitz, K.} \&
  \bibinfo{author}{Tenenbaum, J.~B.}
\newblock \bibinfo{title}{{Church: a langauge for generative models}}.
\newblock In \emph{\bibinfo{booktitle}{Uncertainty in Artificial Intelligence}}
  (\bibinfo{year}{2008}).

\bibitem{SalHinton07}
\bibinfo{author}{Salakhutdinov, R.} \& \bibinfo{author}{Hinton, G.}
\newblock \bibinfo{title}{Deep {B}oltzmann machines}.
\newblock In \emph{\bibinfo{booktitle}{Proceedings of the International
  Conference on Artificial Intelligence and Statistics}},
  vol.~\bibinfo{volume}{5}.

\bibitem{pearlprobabilistic}
\bibinfo{author}{Pearl, J.}
\newblock \emph{\bibinfo{title}{Probabilistic Reasoning in Intelligent Systems:
  Networks of Plausible Inference}} (\bibinfo{publisher}{Morgan Kaufmann
  Publishers}, \bibinfo{address}{San Francisco}, \bibinfo{year}{1988}).

\bibitem{linhigh}
\bibinfo{author}{Lin, M.}, \bibinfo{author}{Lebedev, I.} \&
  \bibinfo{author}{Wawrzynek, J.}
\newblock \bibinfo{title}{High-throughput bayesian computing machine with
  reconfigurable hardware}.
\newblock In \emph{\bibinfo{booktitle}{Proceedings of the 18th annual ACM/SIGDA
  international symposium on Field programmable gate arrays}},
  \bibinfo{pages}{73--82} (\bibinfo{organization}{ACM}, \bibinfo{year}{2010}).

\bibitem{vigodacontinuous}
\bibinfo{author}{Vigoda, B.~W.}
\newblock \emph{\bibinfo{title}{Continuous-time analog circuits for statistical
  signal processing}}.
\newblock Ph.D. thesis, \bibinfo{school}{Massachusetts Institute of Technology}
  (\bibinfo{year}{2003}).

\bibitem{shannonsymbolic}
\bibinfo{author}{Shannon, C.~E.}
\newblock \emph{\bibinfo{title}{A symbolic analysis of relay and switching
  circuits}}.
\newblock Ph.D. thesis, \bibinfo{school}{Massachusetts Institute of Technology}
  (\bibinfo{year}{1940}).

\bibitem{VMEJ-circuits-supplemental}
\bibinfo{author}{Mansinghka, V.} \& \bibinfo{author}{Jonas, E.}
\newblock \bibinfo{title}{Supplementary material on stochastic digital
  circuits}  (\bibinfo{year}{2014}).
\newblock
  \urlprefix\url{http://probcomp.csail.mit.edu/VMEJ-circuits-supplement.pdf}.

\bibitem{metropolisequation}
\bibinfo{author}{Metropolis, N.}, \bibinfo{author}{Rosenbluth, A.~W.},
  \bibinfo{author}{Rosenbluth, M.~N.}, \bibinfo{author}{Teller, A.~H.} \&
  \bibinfo{author}{Teller, E.}
\newblock \bibinfo{title}{Equation of state calculations by fast computing
  machines}.
\newblock \emph{\bibinfo{journal}{The journal of chemical physics}}
  \textbf{\bibinfo{volume}{21}}, \bibinfo{pages}{1087} (\bibinfo{year}{1953}).

\bibitem{gemanstochastic}
\bibinfo{author}{Geman, S.} \& \bibinfo{author}{Geman, D.}
\newblock \bibinfo{title}{Stochastic relaxation, gibbs distributions, and the
  bayesian restoration of images}.
\newblock \emph{\bibinfo{journal}{Pattern Analysis and Machine Intelligence,
  IEEE Transactions on}} \bibinfo{pages}{721--741} (\bibinfo{year}{1984}).

\bibitem{andrieuintroduction}
\bibinfo{author}{Andrieu, C.}, \bibinfo{author}{De~Freitas, N.},
  \bibinfo{author}{Doucet, A.} \& \bibinfo{author}{Jordan, M.~I.}
\newblock \bibinfo{title}{An introduction to mcmc for machine learning}.
\newblock \emph{\bibinfo{journal}{Machine learning}}
  \textbf{\bibinfo{volume}{50}}, \bibinfo{pages}{5--43} (\bibinfo{year}{2003}).

\bibitem{ward1990computation}
\bibinfo{author}{Ward~Jr, S.~A.} \& \bibinfo{author}{Halstead, R.~H.}
\newblock \emph{\bibinfo{title}{Computation Structures.}}
  (\bibinfo{publisher}{The MIT press}, \bibinfo{year}{1990}).

\bibitem{marsagliaxorshift}
\bibinfo{author}{Marsaglia, G.}
\newblock \bibinfo{title}{Xorshift rngs}.
\newblock \emph{\bibinfo{journal}{Journal of Statistical Software}}
  \textbf{\bibinfo{volume}{8}}, \bibinfo{pages}{1--6} (\bibinfo{year}{2003}).

\bibitem{bertrillions}
\bibinfo{author}{Wang, F.} \& \bibinfo{author}{Agrawal, V.~D.}
\newblock \bibinfo{title}{Soft error rate determination for nanometer cmos vlsi
  logic}.
\newblock In \emph{\bibinfo{booktitle}{40th Southwest Symposium on Systems
  Theory}}, \bibinfo{pages}{324--328} (\bibinfo{year}{2008}).

\bibitem{marrcooperative}
\bibinfo{author}{Marr, D.} \& \bibinfo{author}{Poggio, T.}
\newblock \bibinfo{title}{Cooperative computation of stereo disparity}.
\newblock \emph{\bibinfo{journal}{Science}} \textbf{\bibinfo{volume}{194}},
  \bibinfo{pages}{283--287} (\bibinfo{year}{1976}).

\bibitem{szeliskicomparative}
\bibinfo{author}{Szeliski, R.} \emph{et~al.}
\newblock \bibinfo{title}{A comparative study of energy minimization methods
  for markov random fields with smoothness-based priors}.
\newblock \emph{\bibinfo{journal}{Pattern Analysis and Machine Intelligence,
  IEEE Transactions on}} \textbf{\bibinfo{volume}{30}},
  \bibinfo{pages}{1068--1080} (\bibinfo{year}{2008}).

\bibitem{fergusonbayesian}
\bibinfo{author}{Ferguson, T.~S.}
\newblock \bibinfo{title}{A bayesian analysis of some nonparametric problems}.
\newblock \emph{\bibinfo{journal}{The annals of statistics}}
  \bibinfo{pages}{209--230} (\bibinfo{year}{1973}).

\bibitem{rasmusseninfinite}
\bibinfo{author}{Rasmussen, C.~E.}
\newblock \bibinfo{title}{The infinite gaussian mixture model}.
\newblock \emph{\bibinfo{journal}{Advances in neural information processing
  systems}} \textbf{\bibinfo{volume}{12}}, \bibinfo{pages}{2}
  (\bibinfo{year}{2000}).

\bibitem{andersonrational}
\bibinfo{author}{Anderson, J.~R.} \& \bibinfo{author}{Matessa, M.}
\newblock \bibinfo{title}{A rational analysis of categorization}.
\newblock In \emph{\bibinfo{booktitle}{Proc. of 7th International Machine
  Learning Conference}}, \bibinfo{pages}{76--84} (\bibinfo{year}{1990}).

\bibitem{griffithscategorization}
\bibinfo{author}{Griffiths, T.~L.}, \bibinfo{author}{Sanborn, A.~N.},
  \bibinfo{author}{Canini, K.~R.} \& \bibinfo{author}{Navarro, D.~J.}
\newblock \bibinfo{title}{Categorization as nonparametric bayesian density
  estimation}.
\newblock \emph{\bibinfo{journal}{The probabilistic mind: Prospects for
  Bayesian cognitive science}} \bibinfo{pages}{303--328}
  (\bibinfo{year}{2008}).

\bibitem{spintronicnand}
\bibinfo{author}{Imre, A.} \emph{et~al.}
\newblock \bibinfo{title}{Majority logic gate for magnetic quantum-dot cellular
  automata}.
\newblock \emph{\bibinfo{journal}{Science}} \textbf{\bibinfo{volume}{311}},
  \bibinfo{pages}{205--208} (\bibinfo{year}{2006}).

\bibitem{kschischangfactor}
\bibinfo{author}{Kschischang, F.~R.}, \bibinfo{author}{Frey, B.~J.} \&
  \bibinfo{author}{Loeliger, H.-A.}
\newblock \bibinfo{title}{Factor graphs and the sum-product algorithm}.
\newblock \emph{\bibinfo{journal}{Information Theory, IEEE Transactions on}}
  \textbf{\bibinfo{volume}{47}}, \bibinfo{pages}{498--519}
  (\bibinfo{year}{2001}).

\bibitem{fiser2010statistically}
\bibinfo{author}{Fiser, J.}, \bibinfo{author}{Berkes, P.},
  \bibinfo{author}{Orb{\'a}n, G.} \& \bibinfo{author}{Lengyel, M.}
\newblock \bibinfo{title}{Statistically optimal perception and learning: from
  behavior to neural representations}.
\newblock \emph{\bibinfo{journal}{Trends in cognitive sciences}}
  \textbf{\bibinfo{volume}{14}}, \bibinfo{pages}{119--130}
  (\bibinfo{year}{2010}).

\bibitem{berkes2011spontaneous}
\bibinfo{author}{Berkes, P.}, \bibinfo{author}{Orb{\'a}n, G.},
  \bibinfo{author}{Lengyel, M.} \& \bibinfo{author}{Fiser, J.}
\newblock \bibinfo{title}{Spontaneous cortical activity reveals hallmarks of an
  optimal internal model of the environment}.
\newblock \emph{\bibinfo{journal}{Science}} \textbf{\bibinfo{volume}{331}},
  \bibinfo{pages}{83--87} (\bibinfo{year}{2011}).

\bibitem{probbrains}
\bibinfo{author}{Pouget, A.}, \bibinfo{author}{Beck, J.}, \bibinfo{author}{Ma,
  W.~J.} \& \bibinfo{author}{Latham, P.~E.}
\newblock \bibinfo{title}{Probabilistic brains: knowns and unknowns}.
\newblock \emph{\bibinfo{journal}{Nature Neuroscience}}
  \textbf{\bibinfo{volume}{16}}, \bibinfo{pages}{1170--1178}
  (\bibinfo{year}{2013}).

\bibitem{diaconismarkov}
\bibinfo{author}{Diaconis, P.}
\newblock \bibinfo{title}{The markov chain monte carlo revolution}.
\newblock \emph{\bibinfo{journal}{Bulletin of the American Mathematical
  Society}} \textbf{\bibinfo{volume}{46}}, \bibinfo{pages}{179--205}
  (\bibinfo{year}{2009}).

\bibitem{dagum1997optimal}
\bibinfo{author}{Dagum, P.} \& \bibinfo{author}{Luby, M.}
\newblock \bibinfo{title}{An optimal approximation algorithm for bayesian
  inference}.
\newblock \emph{\bibinfo{journal}{Artificial Intelligence}}
  \textbf{\bibinfo{volume}{93}}, \bibinfo{pages}{1--27} (\bibinfo{year}{1997}).

\bibitem{intelInstructionErrors}
\bibinfo{author}{Weaver, C.}, \bibinfo{author}{Emer, J.},
  \bibinfo{author}{Mukherjee, S.} \& \bibinfo{author}{Reinhardt, S.}
\newblock \bibinfo{title}{Techniques to reduce the soft error rate of a
  high-performance microprocessor}.
\newblock In \emph{\bibinfo{booktitle}{Computer Architecture, 2004.
  Proceedings. 31st Annual International Symposium on}},
  \bibinfo{pages}{264--275} (\bibinfo{year}{2004}).

\bibitem{shepardnoise}
\bibinfo{author}{Shepard, K.~L.} \& \bibinfo{author}{Narayanan, V.}
\newblock \bibinfo{title}{Noise in deep submicron digital design}.
\newblock In \emph{\bibinfo{booktitle}{Proceedings of the 1996 IEEE/ACM
  international conference on Computer-aided design}},
  \bibinfo{pages}{524--531} (\bibinfo{organization}{IEEE Computer Society},
  \bibinfo{year}{1997}).

\bibitem{elowitzsynthetic}
\bibinfo{author}{Elowitz, M.~B.} \& \bibinfo{author}{Leibler, S.}
\newblock \bibinfo{title}{A synthetic oscillatory network of transcriptional
  regulators}.
\newblock \emph{\bibinfo{journal}{Nature}} \textbf{\bibinfo{volume}{403}},
  \bibinfo{pages}{335--338} (\bibinfo{year}{2000}).

\bibitem{barroso2007case}
\bibinfo{author}{Barroso, L.~A.} \& \bibinfo{author}{Holzle, U.}
\newblock \bibinfo{title}{The case for energy-proportional computing}.
\newblock \emph{\bibinfo{journal}{Computer}} \textbf{\bibinfo{volume}{40}},
  \bibinfo{pages}{33--37} (\bibinfo{year}{2007}).

\bibitem{flinn1999energy}
\bibinfo{author}{Flinn, J.} \& \bibinfo{author}{Satyanarayanan, M.}
\newblock \bibinfo{title}{Energy-aware adaptation for mobile applications}.
\newblock \emph{\bibinfo{journal}{ACM SIGOPS Operating Systems Review}}
  \textbf{\bibinfo{volume}{33}}, \bibinfo{pages}{48--63}
  (\bibinfo{year}{1999}).

\bibitem{mcadams1997stochastic}
\bibinfo{author}{McAdams, H.~H.} \& \bibinfo{author}{Arkin, A.}
\newblock \bibinfo{title}{Stochastic mechanisms in gene expression}.
\newblock \emph{\bibinfo{journal}{Proceedings of the National Academy of
  Sciences}} \textbf{\bibinfo{volume}{94}}, \bibinfo{pages}{814--819}
  (\bibinfo{year}{1997}).

\bibitem{lecun1998mnist}
\bibinfo{author}{LeCun, Y.} \& \bibinfo{author}{Cortes, C.}
\newblock \bibinfo{title}{The mnist database of handwritten digits}
  (\bibinfo{year}{1998}).

\end{thebibliography}

%}

%%
%% TABLES
%%
%% If there are any tables, put them here.
%%

\end{document}